# Automated Bridge Component Recognition using Video Data

Yasutaka Narazaki[a], Vedhus Hoskere[a], Tu A. Hoang[a], Billie F. Spencer Jr.[a]

[a] Department of Civil and Environmental Engineering,
University of Illinois at Urbana-Champaign, Urbana-Champaign, USA.

## Abstract

This paper investigates the automated recognition of structural bridge components using video data. Although understanding video data for structural inspections is straightforward for human inspectors, the implementation of the same task using machine learning methods has not been fully realized. In particular, single-frame image processing techniques, such as convolutional neural networks (CNNs), are not expected to identify structural components accurately when the image is a close-up view, lacking contextual information regarding where on the structure the image originates. Inspired by the significant progress in video processing techniques, this study investigates automated bridge component recognition using video data, where the information from the past frames is used to augment the understanding of the current frame. A new simulated video dataset is created to train the machine learning algorithms. Then, convolutional Neural Networks (CNNs) with recurrent architectures are designed and applied to implement the automated bridge component recognition task. Results are presented for simulated video data.

## 1   Introduction

Bridges are critical parts of transportation infrastructure that need to be maintained appropriately through proper inspection to ensure safe operation. To support the time-consuming and labor-intensive visual inspections, automated image processing techniques have been applied to still images of bridges or their structural components (local features [1]–[3], methods based on convolutional neural networks [4]–[10], etc.). Promising results have been obtained for automated recognition of critical bridge components or damage on the component surfaces. In such methods, damage recognition is most likely to be successful when the image is a close-up view of the component surfaces, while the bridge component recognition needs global cues from the entire structure. During bridge inspection by humans, this trade-off is easily resolved by first examining the entire structure, and then moving close to the structural components of interest. However, the implementation of the visual recognition task during the bridge inspection process is not straightforward, because the naïve application of convolutional neural networks (CNNs) and their variants processes each frame of the video independently without leveraging information from previous frames.

Neural networks with recurrent architectures have been proposed as effective methods for modeling a sequence from collected data (a video is a sequence of images). Simple recurrent neural networks (RNNs) can be implemented by regarding the past input data as additional channels of the current input data and optimizing the parameters by backpropagation through time [11]. MaskTrack Convnet [12] is a CNN architecture similar in concept to the simple RNN, where to track and segment the object, an input image is augmented by the estimated object mask from the previous step.

Despite the conceptual simplicity of such RNN architectures, learning from a sequence of data becomes difficult as the length of the data increases. Gradient-based parameter updating for such sequences is known to be inefficient, because backpropagation error vanishes or explodes rapidly (i.e., vanishing gradient problem) [13]. Long short-term memory (LSTM) units [13] have been proposed to circumvent the problem by explicitly implementing memory in the architecture (i.e., constant error carrousel). The data and error flow from/into memory is controlled by "gates", which are modeled by sigmoid functions. Gated recurrent units (GRU) have also been proposed as a simplified alternative to the LSTM unit [14].

The LSTM and GRU cells have been integrated into CNN architectures to model sequences of 2D maps, including images and grid measurement data. Convolutional Gated Recurrent Networks [15] were proposed to improve the semantic segmentation and object tracking tasks for the video stream data of urban scenes. Convolutional LSTM (ConvLSTM) network [16] were proposed to perform weather forecasting from recent radar echo sequences represented by 2D maps. These methods have the potential to address the lack of global information in close-up images that plagues the task of bridge component recognition required for automated structural inspection.

This study investigates the task of automated bridge component recognition by combining pre-trained single image-based processing method with recurrent units to estimate the bridge component labels. First, pixel-wise bridge component labels (semantic segmentation) are estimated using a single frame-based deep fully convolutional networks (FCNs) [17] with ResNet connections [18]. A bridge component classification dataset [5][6] is used to learn visual features of the structures of a variety of bridge types. Then, simple RNN units and ConvLSTM units are added to the FCN architecture to introduce recurrence, while maintaining computational efficiency. To train the recurrent parts of the network, a video dataset is created, where a UAV-like agent navigates randomly in a virtual world and records videos and corresponding ground truth bridge component recognitions. Results for the automated bridge component recognition task are presented for simulated video data, as well as video collected in the field.

## 2 Method

This section discusses the problem of recognizing the bridge components from video data. Single frame-based semantic segmentation using fully convolutional networks (FCNs) are discussed first, and two methods of introducing recurrence to the networks – simple RNN and LSTM are discussed. These methods are combined to perform the task of bridge component recognition, while keeping the complexity of the problem within an acceptable range.

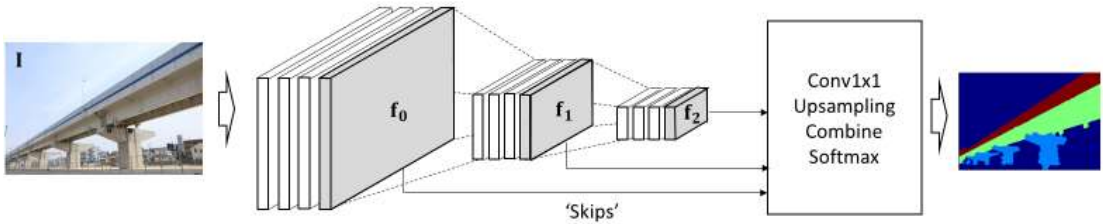

Figure 1. Illustration of the fully convolutional network (FCN) architecture.

## 2.1. Fully convolutional networks (FCNs)

Fully convolutional networks [17] are proposed as an effective method to extend the normal convolutional neural networks (CNNs) to perform pixel-wise labelling (semantic segmentation) tasks (Figure 1). Similar to normal CNNs, the input image to the FCNs is passed through non-linear convolutional layers and max-pooling layers. In contrast to normal CNNs which assumes the last layer output ($\mathbf{f_2}$ in Figure 1) as a single feature vector representing the input image, the FCNs interpret the same $\mathbf{f_2}$ as a (down-sampled) feature map, which stores feature vectors at the corresponding locations of the image. Therefore, the label at each element location of the down-sampled feature map $\mathbf{f_2}$ can be estimated by classifying each feature vector into an appropriate class. This estimation step can be implemented by a convolution with filter size $1 \times 1$.

Generally, the estimated labels from the down-sampled feature map does not have enough spatial resolution, because the max-pooling layers reduce the spatial information. In FCNs, to estimate label maps with higher spatial resolutions, output from multiple layers with different resolutions are "skipped" and merged with the estimation results. In the architecture in Figure 1, the estimated label map (before applying softmax scaling) is up-sampled to the resolution of $\mathbf{f_1}$, and added to the (unscaled) estimated label map from the layer $\mathbf{f_1}$. The combined estimated label map is then up-sampled to the resolution of $\mathbf{f_0}$ and added to the estimated feature map from the layer $\mathbf{f_0}$. Finally, the merged label map is up-sampled to the original resolution of the image. The up-sampling filters are either fixed or learned, except for the last up-sampling operation, where the up-sampling filter is fixed to the bilinear interpolation filter. The FCN architectures used with the famous CNN architectures (e.g., VGG architectures [19]) have been frequently applied to semantic segmentation problem of objects (e.g., [17], [20]–[22]).

## 2.2. Recurrent Neural Networks (RNNs)

Recurrent neural networks refer to neural networks with feedbacks. As shown in the left side of Figure 2 [23], a recurrent unit takes both the input data $x$ and the output of the unit at previous time step, and apply nonlinear operations to compute the output at the current time step. The recurrent architecture can be "unfolded" to create an equivalent graph which can be regarded as a deep architecture with shared parameters $W$ (Figure 2 right).

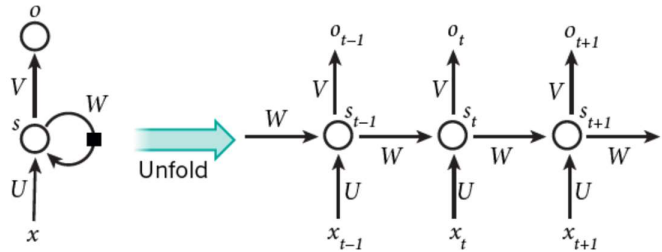

Figure 2. Illustration of Recurrent Neural Networks (RNNs). Quote from [23].

The simple recurrent unit can be created by implementing normal matrix multiplication or convolution, followed by a nonlinear activation function, i.e.

$$o_t = f(U \cdot x + W \cdot s_{t-1}) \quad \text{or} \quad o_t = f(U * x + W * s_{t-1}) \tag{1}$$

where $\cdot$ denotes matrix product and $*$ denotes convolution. The recurrent networks thus created can be trained by gradient descent algorithms (backpropagation through time [11]).

A problem of the simple RNN is the difficulty of learning patterns of long sequences, known as the vanishing gradient problem [13], [23]. As the error gradient propagates backward in time domain (see Figure 2 right), the gradient explodes or vanishes rapidly, which makes the learning of long-term patterns impractical. The vanishing gradient problem is particularly problematic in this study, because the understanding of the structure obtained at a certain global to semi-global view needs to affect the later structural component recognition while the viewer takes a close look at the structural components.

The Long Short-Term Memory (LSTM) is a recurrent unit designed to circumvent the vanishing gradient problem [13]. The structure of the LSTM cell is illustrated in Figure 3. Following Figure 2, the input to the cell at time $t$ is linearly transformed by appropriate weights and passed through a nonlinear activation function $g$ (e.g., the tanh function). Then, the output of $g$ is multiplied by the value of an "input gate". The input gate is modelled by a sigmoid function of the input $x_t$ and the previous output $o_{t-1}$. When the sigmoid function takes the value close to one, the input signal flows into the cell and added to the hidden state $s$. The hidden state $s$ is kept constant in time except for the addition of input signal (this part is called "constant error carrousel"). The hidden state is passed through a nonlinear function $h$ (e.g., tanh function) and again multiplied by the value of an "output gate" to control if the hidden state can affect the output of the network. The advantage of explicitly implementing the memory by the constant error carrousel is that the gradient does not vanish or explodes during training (mathematical proof is provided in [13]).

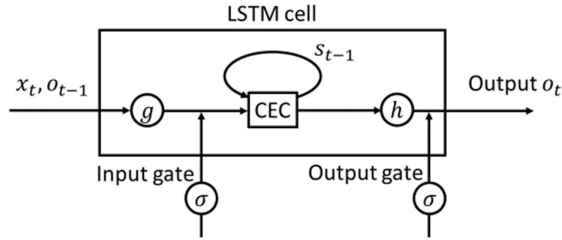

Figure 3. Illustration of the Long Short-TermMemory (LSTM) cell.

Recurrent neural networks including LSTM have been applied to the modelling of video data (sequences of images). Parazzi, et al. [12] shows that stacking current input image and estimated mask at the previous time step and feeding the augmented input into convolutional layers are effective steps to track an object in the video. Shi, et al. [16] developed a convolutional LSTM (ConvLSTM) architecture, where the equations for the units are expressed as follows:

$$s_{t+1} = \sigma(W_{xf} * x_{t+1} + W_{of} * o_t + W_{sf} \circ s_t + b_f) \circ s_t \\ + \sigma(W_{xi} * x_{t+1} + W_{oi} * o_t + W_{si} \circ s_t + b_i) \\ \circ \tanh(W_{xs} * x_{t+1} + W_{os} * o_t + b_s) \quad (2)$$

$$o_t^j = \sigma(W_{xo} * x_{t+1} + W_{oo} * o_t + W_{so} \circ s_t + b_o) \circ \tanh(s_t) \quad (3)$$

where * denotes convolution and ∘ denotes element-wise product. The first equation shows the update of the hidden states, where the input to the cell is gated by a sigmoid function. Also, an additional "forget gate" is implemented in this model by multiplying a sigmoid function to the previous state. The second equation shows the output of the ConvLSTM cell expressed by a product of the output from the CEC and the output gate. Siam, et al. [15] used similar recurrent unit (Gated Recurrent Unit [14]) with FCNs to get improved semantic segmentation of video data.

### 2.3. Bridge component recognition using pre-trained FCN and additional recurrent architectures

The network architecture used in this study is illustrated in Figure 4. First, a deep single image-based FCN is applied to extract a map of label predictions (before scaling by softmax). Then, additional RNN layers are added after the lowest resolution prediction layer (prediction from $f_2$ in the example of Figure 4). Finally, the output from the RNN layers and other skipped layers with higher resolutions are combined to generate the final estimated label maps.

Compared with fully integrated CNN-RNN architectures [14], [16], the FCN and RNN can be trained separately using different datasets. This feature is particularly advantageous for this study, because the dataset for single image-based FCN includes a variety of bridge types, while the video dataset includes simulated video records during random navigation around a single bridge. If the CNN architecture with recurrent units are trained end-to-end using the video dataset, the resulting network is expected to show overfitting.

Furthermore, the RNN units are inserted only after the lowest resolution prediction layer, because the RNN units in this study are used to memorize where the video is focused, rather than improving the level of details of the estimated map. In addition to the reduction in the size of the problem, this architecture is advantageous because predictions of skipped layers can be pre-computed, and the RNN units can be trained without repeating the FCN computation.

Two types of RNN units are tested in this study – simple RNN and ConvLSTM units. For the simple RNN units, the input to the unit is augmented by the output of the unit at the previous time step, and the convolution with ReLU activation function is applied in the unit. Alternatively, ConvLSTM units are inserted into the RNN of the architecture and the effectiveness for modelling long-term patterns are evaluated.

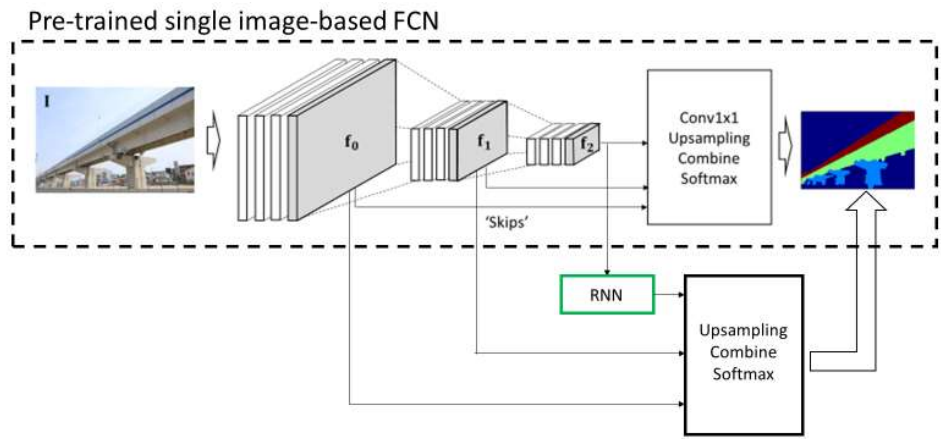

Figure 4. Illustration of the network architecture used in this study.

## 3. Datasets

### 3.1. Dataset for single image-based bridge component recognition

Two datasets were collected by combining existing datasets and newly-labeled datasets. The first dataset, termed the scene classification dataset, contains 11,897 outdoor scene images. The pixel-wise labels of the existing datasets are transferred to 10 high-level scene classes (building, greenery, person, pavement, signs and poles, vehicles, bridges, water, sky, and others).

The second dataset, termed the bridge component classification dataset, contains 1,563 bridge images with pixel-wise bridge component labels of 5 classes: Non-bridge, Columns,

Beams & Slabs, Other Structural, and Other Nonstructural. Both datasets are resized, such that the longer dimension of the image has 320 pixels. The details of the scene classification dataset and the bridge component recognition datasets are provided in [6].

### 3.2. Video dataset for RNN training

A new simulated video dataset imitating random navigation of a UAV around a concrete girder bridge was created for this study using the Unity3D game engine [24]. The steps to create the dataset are similar to the steps to create the SYNTHIA dataset [21]. However, this dataset navigates in 3D space with more abrupt changes of the heading, pitch, and altitude. The resolution of the video was set to 240 × 320, and 37,081 training images and 2000 testing images are generated for this study.

The example frames of the video are shown in Figure 7. Labels follow the rules for the bridge component classification dataset for single image processing. The labels do not have "other structural" class, because no such component exists in this bridge. The depth map is also retrieved, although the data is not used for this study.

## 4. Training and Results

### 4.1. Fully convolutional networks

A 45-layer FCN with residual network connections [18], batch normalization [25], median frequency balancing [26], and weight decay [27] is designed for this study (see the Appendix for the details). Following [6], two FCNs are trained – one for scene classification trained using the scene classification dataset, and another FCN concatenated sequentially to estimate bridge component labels from both the input image and the estimated scene labels. This architecture has been shown to be effective at producing a reduced number of false-positive detections. The details of the training (data augmentation, learning rate, etc.) followed the steps in [6].

The testing results of the trained FCN on the bridge classification dataset (test set) is shown in Figure 5(a). The total pixel-wise accuracy is 82.30%, which validates the recognition capabilities of the trained classifier. In contrast, the bridge component recognition results of the same FCN evaluated on the test set of the video dataset are unsatisfactory. The confusion matrix in Figure 5(b) shows much lower accuracy for Beams & Slabs class and Other (Nonstructural) classes. Although the accuracy for the column class appears to be improved at a first glance, the comparison is not straightforward, because the bridge component classification dataset contains a variety of bridge types, while the video dataset includes images of a single concrete girder bridge only. The total pixel-wise accuracy for the video dataset is 65.0%, which shows the difficulty in recognizing bridge components from a single frame of the video when the video captures close-up views as well as global views. Examples of estimated labels are shown in Figure 7 (second column).

### 4.2. Recurrent Neural Networks

As discussed in Section 2, recurrence is introduced to the network by adding two types of recurrent units to the pre-trained single image-based FCN – simple RNN and ConvLSTM. For both cases, 3 layers of the recurrent units with the filter size of 5x5 and the depth of 15 are placed after the lowest resolution prediction map (5x5x5x15 – 15x5x5x15 – 15x5x5x15), followed by a normal convolutional layer (15x1x1x5) to compute the updated prediction. Tensorflow implementation of the ConvLSTM [28] is used, and the RNN parameters are tuned by the Adam optimizer [29]. During training, batch size is set to 1 for both cases, and the ConvLSTM units are unrolled up to 5 time steps. The learning rate is set to $1.0 \times 10^{-4}$ for the first 10 epochs, $1.0 \times 10^{-5}$ for the next 5 epochs, and $1.0 \times 10^{-6}$ for the last 1 epoch (an epoch refers to a set of iterations from the beginning to the end of the training data set). To prevent overfitting of the temporal characteristics of the image sequences, frames are randomly sampled as follows: (i) divide the training data into blocks of size 1,000, (ii) randomly sample 1,000 integers between 0 and 999, and (iii) only use frames of the data block indexed by the integers appeared at least once during the second step.

The confusion matrices for the two recurrent architectures are shown in Figure 6. Total pixel-wise accuracy for the simple RNN and the ConvLSTM units are 74.9% and 80.5%, respectively. Compared to the confusion matrix of the FCN, the effectiveness of the recurrent unit is clearly observed. Moreover, the ConvLSTM outperforms the simple RNN, except for the accuracy for the Beams & Slabs class. Example results in Figure 7 shows the effectiveness of the recurrent units when the FCN fails to recognize the bridge components correctly. Based on the results and discussion in this section, the ConvLSTM units combined with the pre-trained FCN is an effective approach to perform automated bridge component recognition, even if the visual cues of the global structures are temporally unavailable.

## 5. Conclusions

This study investigated the use of recurrent neural networks (RNNs) with a pre-trained fully convolutional network (FCN) to perform the automated bridge component recognition from video data. The bridge component recognition task is not straightforward to solve, because the visual cues of the global structures are lost as the inspector approaches the component. To improve the recognition performance of the FCN using a single image, recurrent units are added to the FCN. By putting the recurrent units only after the lowest resolution prediction layer and training the recurrent unit independently, the RNN parameters were learned in a reasonable amount of time. The architecture with recurrent units outperformed the FCN both quantitatively (pixel-wise accuracy) and qualitatively (example estimated label maps). Moreover, the ConvLSTM units performed significantly better than the simple RNN when the FCN failed to recognize the bridge components. In the future, computation time will need to be thoroughly evaluated to apply the method in near real-time.

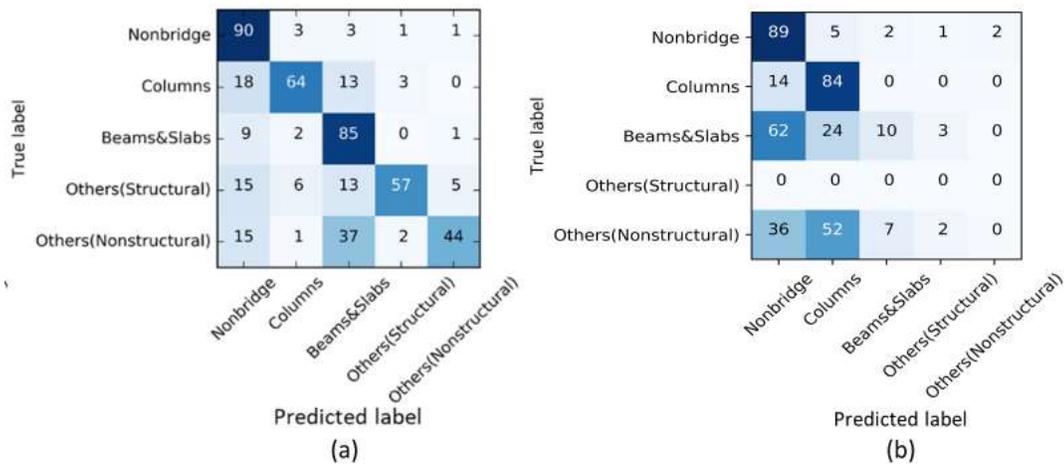

Figure 5 FCN test results (a)Bridge component classification dataset, (b)Video dataset

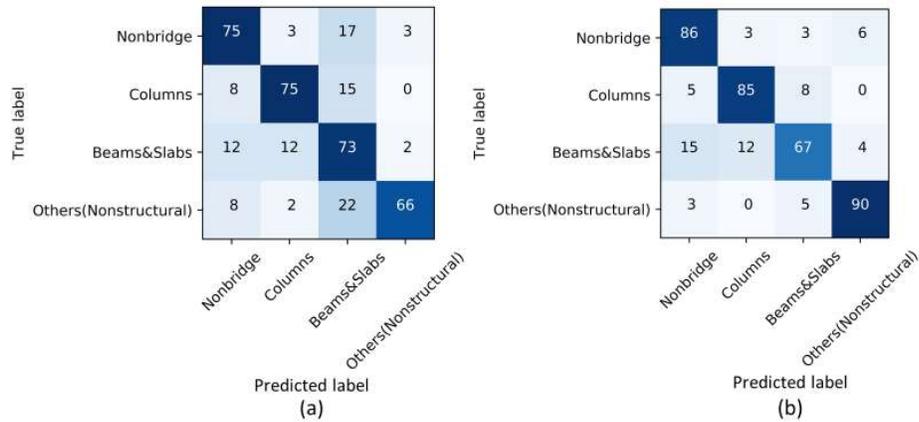

Figure 6. Confusion matrices (a) SimpleRNN, (b) ConvLSTM.

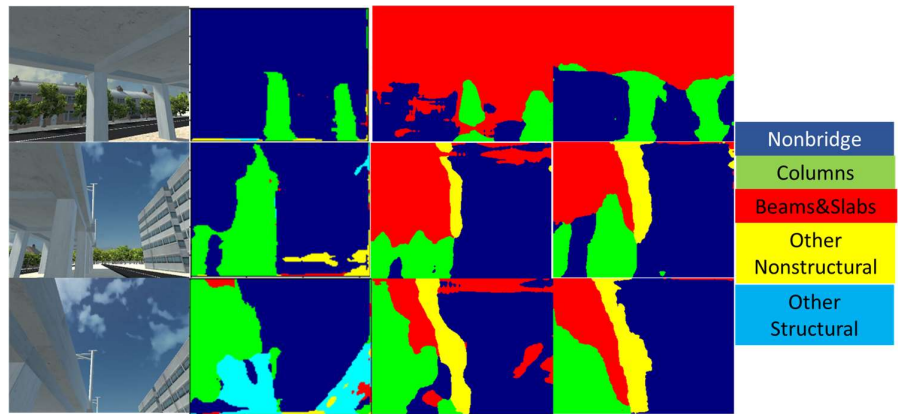

Figure 7. Example results (From left to right: Input image, FCN, FCN-SimpleRNN, FCN-ConvLSTM);

## Appendix – FCN architecture

| \multicolumn{6}{c}{*FCN45* architecture} | | | | | |
|---|---|---|---|---|---|
| Name | Filt. Size | ResNet connect. | Name | Filt. Size | ResNet connect. |
| Conv0 | 7x7x64 (stride 2) |  | Conv22 | 3x3x128 | Maxpool1 |
| Conv1 | 3x3x64 |  | Conv23 | 3x3x128 |  |
| Conv2 | 3x3x64 | Conv0 | Conv24 | 3x3x128 | Conv22 |
| Conv3 | 3x3x64 |  | Conv25 | 3x3x128 |  |
| Conv4 | 3x3x64 | Conv2 | Conv26 | 3x3x128 | Conv24 |
| Conv5 | 3x3x64 |  | Conv27 | 3x3x128 |  |
| Conv6 | 3x3x64 | Conv4 | Conv28 | 3x3x128 | Conv26 |
| Conv7 | 3x3x64 |  | Conv29 | 3x3x128 |  |
| Conv8 | 3x3x64 | Conv6 | Conv30 | 3x3x128 | Conv28 |
| Maxpool0 | 2x2 |  | Conv31 | 3x3x128 |  |
| Conv9 | 3x3x128 |  | Conv32 | 3x3x128 | Conv30 |
| Conv10 | 3x3x128 | Maxpool0 | Maxpool3 | 2x2 |  |
| Conv11 | 3x3x128 |  | Conv33 | 3x3x128 |  |
| Conv12 | 3x3x128 | Conv10 | Conv34 | 3x3x128 | Maxpool2 |
| Conv13 | 3x3x128 |  | Conv35 | 3x3x128 |  |
| Conv14 | 3x3x128 | Conv12 | Conv36 | 3x3x128 | Conv34 |
| Conv15 | 3x3x128 |  | Conv37 | 3x3x128 |  |
| Conv16 | 3x3x128 | Conv14 | Conv38 | 3x3x128 | Conv36 |
| Conv17 | 3x3x128 |  | Conv39 | 3x3x128 |  |
| Conv18 | 3x3x128 | Conv16 | Conv40 | 3x3x128 | Conv38 |
| Conv19 | 3x3x128 |  | Conv41 | 3x3x128 |  |
| Conv20 | 3x3x128 | Conv18 | Conv42 | 3x3x128 | Conv40 |
| Maxpool1 | 2x2 |  | Conv43 | 3x3x128 |  |
| Conv21 | 3x3x128 |  | Conv44 | 3x3x128 | Conv42 |
| Pred. layer | \multicolumn{5}{c}{Single layer FCL} | | | | | |
| # scales | \multicolumn{5}{c}{1} | | | | | |
| Batch size | \multicolumn{5}{c}{10} | | | | | |
| Wt. decay | \multicolumn{5}{c}{0.0001} | | | | | |
| Skips | \multicolumn{5}{c}{Conv20, Conv32} | | | | | |